\pdfoutput=1

\documentclass[11pt]{article}

\usepackage[final]{acl}
\usepackage{xcolor}

\usepackage{times}
\usepackage{latexsym}

\usepackage[T1]{fontenc}

\usepackage[utf8]{inputenc}

\usepackage{microtype}

\usepackage{inconsolata}

\usepackage{graphicx}

\usepackage{contour}
\usepackage{booktabs}

\newcommand{\barec}{\textbf{\textsc{Barec}}}

\newcommand{\hide}[1]{}

\newcommand{\TAMARBUTA}{{$\hbar$}}

\newcommand{\DHA}{{\dh}}

\usepackage{arabtex}

\usepackage{utf8}
\usepackage{epstopdf}

\usepackage{multirow}

\newcommand{\BlevAlif}{{\bf 1-alif}}
\newcommand{\BlevBa}{{\bf 2-ba}}
\newcommand{\BlevJim}{{\bf 3-jim}}
\newcommand{\BlevDal}{{\bf 4-dal}}
\newcommand{\BlevHe}{{\bf 5-ha}}
\newcommand{\BlevWaw}{{\bf 6-waw}}
\newcommand{\BlevZay}{{\bf 7-zay}}
\newcommand{\BlevHa}{{\bf 8-\underline{h}a}}
\newcommand{\BlevTa}{{\bf 9-\underline{t}a}}
\newcommand{\BlevYa}{{\bf 10-ya}}
\newcommand{\BlevKaf}{{\bf 11-kaf}}
\newcommand{\BlevLam}{{\bf 12-lam}}
\newcommand{\BlevMim}{{\bf 13-mim}}
\newcommand{\BlevNun}{{\bf 14-nun}}
\newcommand{\BlevSin}{{\bf 15-sin}}
\newcommand{\BlevAyn}{{\bf 16-ayn}}
\newcommand{\BlevFa}{{\bf 17-fa}}
\newcommand{\BlevSad}{{\bf 18-sad}}
\newcommand{\BlevQaf}{{\bf 19-qaf}}

\title{Guidelines for Fine-grained Sentence-level\\ Arabic Readability Annotation}

\author{Nizar Habash,\textsuperscript{\textdagger} Hanada Taha-Thomure,\textsuperscript{$\ddagger$} Khalid N. Elmadani,\textsuperscript{\textdagger}\\  \textbf{Zeina Zeino,}\textsuperscript{$\ddagger$} \textbf{Abdallah Abushmaes}\textsuperscript{\textdagger\textdagger}
  \\
  \textsuperscript{\textdagger}Computational Approaches to Modeling Language Lab, New York University Abu Dhabi\\
   \textsuperscript{$\ddagger$}Zai Arabic Language Research Centre, Zayed University\\
  \textsuperscript{\textdagger}\textsuperscript{\textdagger}Abu Dhabi Arabic Language Centre\\
  \texttt{nizar.habash@nyu.edu, Hanada.Thomure@zu.ac.ae}
  }

\begin{document}
\maketitle

\setcode{utf8}
\vocalize


\begin{abstract}
This paper presents the annotation guidelines of the Balanced Arabic Readability Evaluation Corpus ({\barec}), a large-scale resource for fine-grained sentence-level readability assessment in Arabic. {\barec} includes 69,441 sentences (1M+ words) labeled across 19 levels, from kindergarten to postgraduate. Based on the Taha/Arabi21 framework, the guidelines were refined through iterative training with native Arabic-speaking educators. We highlight key linguistic, pedagogical, and cognitive factors in determining readability and report high inter-annotator agreement: Quadratic Weighted Kappa 81.8\% (substantial/excellent agreement) in the last annotation phase.  We also benchmark automatic readability models across multiple classification granularities (19-, 7-, 5-, and 3-level). The corpus and guidelines are publicly available.\footnote{\label{barec-site}\url{http://barec.camel-lab.com}}


\end{list}
\end{abstract}


 \section{Introduction}

Text readability plays a crucial role in comprehension, retention, reading speed, and engagement \cite{DuBay:2004:principles}. When texts exceed a reader’s ability, they can lead to frustration and disengagement \cite{klare1963measurement}. Readability is shaped by both the content and presentation \cite{Nassiri:2023}. In educational settings, readability leveling is widely used to align texts with students' reading abilities, promoting independent and more effective learning \cite{allington2015research,BarberKlauda2020}.

Fine-grained readability systems, like Fountas and Pinnell’s 27-level scale in English \cite{fountas2006leveled}, and Taha’s 19-level Arabic system \cite{Taha:2017:guidelines}, guide progression from early readers to adult fluency. These levels support instructional goals and can be mapped to broader categories for practical use in NLP.


We present the Balanced Arabic Readability Evaluation Corpus ({\barec}), a large-scale dataset of  {69K+} sentences\footnote{We use \textit{sentence} to refer to syntactic sentences as well as shorter standalone text segments (e.g., phrases or titles).}
(1M+ words) across  a broad space of genres and 19 readability levels. Based on the Taha/Arabi21 framework \cite{Taha:2017:guidelines}, which has been instrumental in tagging over 9,000 children's books, {\barec} guidelines enable standardized, sentence-level readability evaluation across diverse genres and educational levels,
ranging from kindergarten to postgraduate comprehension (see Table~\ref{intro-example}).
Our contributions are as follows:

\begin{table}[t]
\centering
\tabcolsep1pt
\small
\begin{tabular}{cclp{4.9cm}}
\toprule
\textbf{RL} & \textbf{Grade} & \multicolumn{1}{c}{\textbf{Example}}\\
\midrule
\textbf{1}  & \textbf{KG}    & Ball & \multicolumn{1}{r}{\<كُرَة>}\\
\textbf{3}  & \textbf{1st}   & The bedroom & \multicolumn{1}{r}{\<غُرْفَةُ النَّوْمِ> } \\
\textbf{6}  & \textbf{2nd}   & \multicolumn{2}{r}{ \<سُلوكي مَسْؤولِيَّتي>} \\
 & & \multicolumn{2}{l}{My behavior is my responsibility} \\
\textbf{10} & \textbf{4th}   & \multicolumn{2}{r}{\<كانت الحديقة واسعة، تطل على شاطئ النيل،> }\\
 & & \multicolumn{2}{l}{The garden was spacious, overlooking the Nile. }\\ 
\textbf{14} & \textbf{8th}   &\multicolumn{2}{r}{ \<تعريف أصول الفقه> }\\
& & \multicolumn{2}{l}{Definition of Islamic Jurisprudence Principles} \\
\textbf{17} & \textbf{Uni}   & \multicolumn{2}{r}{\<بين طعن القَنا وخَفْق البُنودِ>} \\
& & \multicolumn{2}{l}{Between lance thrusts and ensign flutters}\\
\bottomrule
\end{tabular}
\caption{Examples by Reading Level (RL) and grade. }
\label{intro-example}
\end{table}

\begin{itemize}
\itemsep0pt
   \item We \textbf{define detailed annotation guidelines} for Arabic sentence-level readability across a fine-grained 19-level scale.
   %

  \item We \textbf{apply and refine these guidelines} through annotation of a diverse, large-scale corpus, analyzing annotator agreement and sources of difficulty in this nuanced task.
  
 \item We \textbf{build and evaluate readability models} across multiple granularities (19, 7, 5, and 3 levels) to provide baseline results for various research and application needs.

\end{itemize}



Next, \S\ref{sec:related} reviews related work, \S\ref{sec:guidelines} outlines the annotation framework, \S\ref{sec:corpus} covers data selection, and \S\ref{sec:eval} discusses evaluation results.

 

\begin{table*}[t]
\centering
\setlength{\tabcolsep}{2pt} 
\renewcommand{\arraystretch}{1.1} 
\begin{small}
\begin{tabular}{lllclrl}
\toprule
\textbf{Authors} & \textbf{Project} & \textbf{Metric} & \textbf{Levels} & \textbf{Unit} & \textbf{Size} & \textbf{Content} \\

\midrule
\newcite{Al-Khalifa:2010:automatic}     & Arability            & Readability & 3            & Document & 150     & School Textbooks        \\
\newcite{Forsyth:2014:automatic}        & DLI Corpus           & ILR         & 5 (3)        & Document & 179     & L2 Learner        \\
\newcite{Kilgarriff:2014:corpus-based}  & KELLY                & CEFR        & 6            & Word     & 9,000   & Most Frequent           \\
\newcite{Taha:2017:guidelines}          & Taha/Arabi21         & Readability & 19           & Document & 9,000   & Children's Books        \\
\newcite{al-khalil-etal-2020-large}     & SAMER Lexicon        & Readability & 5            & Word     & 40,000  & General Vocab           \\
\newcite{habash-palfreyman-2022-zaebuc} & ZAEBUC               & CEFR        & 6            & Document & 214     & Prompted Essays         \\
\newcite{naous-etal-2024-readme}        & ReadMe++             & CEFR        & 6            & Sentence & 1,945   & Multi-domain            \\
\newcite{soliman2024creating}           & Arabic Vocab Profile                  & CEFR        & 2   & Word     & 1,200   & L2 Learner  (A1, A2)  \\
\newcite{el-haj-etal-2024-dares}        & DARES                & Grade Level & 12           & Sentence & 13,335  & School Textbooks        \\
\newcite{alhafni-etal-2024-samer}       & SAMER Corpus         & Readability & 3            & Word     & 159,265 & Literature              \\
\newcite{bashendy2024qaes}                  & QAES                 & AES         & 7×5 & Document & 195     & Argumentative Essays    \\
\midrule
Our Work                                & BAREC                & Readability & 19 (7–5–3)    & Sentence &  69,441  & Multi-domain            \\
\bottomrule
\end{tabular}
\end{small}
\caption{Overview of Arabic readability and proficiency-related corpora.}
\label{tab:corpora-overview}
\end{table*}

\section{Related Work}
\label{sec:related}





  

\paragraph{Automatic Readability Assessment}
Automatic readability assessment has been widely studied, resulting in numerous datasets and resources \cite{collins-thompson-callan-2004-language,pitler-nenkova-2008-revisiting,feng-etal-2010-comparison,vajjala-meurers-2012-improving,xu-etal-2015-problems,xia-etal-2016-text,nadeem-ostendorf-2018-estimating,vajjala-lucic-2018-onestopenglish,deutsch-etal-2020-linguistic,lee-etal-2021-pushing}. Early English datasets were often derived from textbooks, as their graded content naturally aligns with readability assessment \cite{vajjala-2022-trends}. However, copyright restrictions and limited digitization have driven researchers to crowdsource readability annotations from online sources \cite{vajjala-meurers-2012-improving,vajjala-lucic-2018-onestopenglish} or leverage CEFR-based L2 assessment exams \cite{xia-etal-2016-text}.

\paragraph{Arabic Readability Efforts}
Arabic readability research has explored text leveling and assessment in multiple frameworks \cite{Nassiri:2023}.

\newcite{Taha:2017:guidelines} proposed a 19-level Arabic text leveling framework for educators, inspired by \newcite{fountas2006leveled} and focused on children’s literature. Targeting full texts (books), particularly for early education, with 11 of the 19 levels covering up to 4th grade, the system supports teachers in matching books to students’ reading abilities. \newcite{Taha:2017:guidelines}'s procedural framework outlines ten qualitative and quantitative criteria: text genre, abstractness of ideas, vocabulary and its proximity to dialects, text authenticity, book production quality, content suitability, sentence structure, illustrations, use of diacritics, and word count. The Arab Thought Foundation adopted this framework under its Arabi21 initiative, which funded the leveling of over 9,000 children’s books.

Other efforts applied CEFR leveling to Arabic, including the KELLY project’s frequency-based word lists, manually annotated corpora such as ZAEBUC \cite{habash-palfreyman-2022-zaebuc} and ReadMe++ \cite{naous-etal-2024-readme}, and vocabulary profiling \cite{soliman2024creating}.
\citet{el-haj-etal-2024-dares} introduced DARES, a readability assessment dataset collected from Saudi school materials.
The SAMER project \cite{al-khalil-etal-2020-large} developed a lexicon with a five-level readability scale, leading to the first manually annotated Arabic parallel corpus for text simplification \cite{alhafni-etal-2024-samer}. 
\newcite{bashendy2024qaes}  presented a corpus of Arabic essays annotated across organization and style traits.


Automated readability assessment in Arabic has evolved from rule-based models using surface features \cite{al2004assessment, Al-Khalifa:2010:automatic} to machine learning approaches with POS, morphology \cite{Forsyth:2014:automatic, Saddiki:2018:feature}, and script features like OSMAN \cite{el-haj-rayson-2016-osman}. Recent work \cite{liberato-etal-2024-strategies} shows strong results with pretrained models on the SAMER corpus.

\paragraph{Our Approach} Building on prior work, we curated the {\barec} corpus across diverse genres and readability levels, manually annotating it at the sentence level using adapted Taha/Arabi21 guidelines \cite{Taha:2017:guidelines}. Sentence-level annotation balances the coarse granularity of document-level labels and the limited context of word-level labels. This allows finer control and more objective assessment of textual variation. Table~\ref{tab:corpora-overview} compares {\barec} with earlier efforts. To our knowledge, {\barec} is the largest and most fine-grained manually annotated Arabic readability resource.


\begin{figure*}[th!]
\centering
 \includegraphics[width=0.96\textwidth]{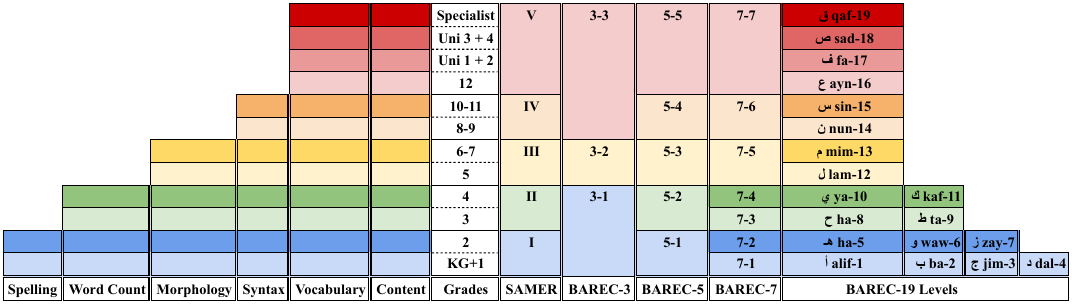}
    \caption{ The {\barec} \textit{Pyramid}  illustrates the relationship across {\barec} levels and linguistic dimensions, three collapsed variants (3 levels, 5 levels and 7 levels), and educational grades.}
\label{fig:barec-pyramid}
\end{figure*}

\section{{\barec}  Annotation Guidelines}
\label{sec:guidelines}

\begin{table*}[h!]
\centering
 \includegraphics[width=0.90\textwidth]{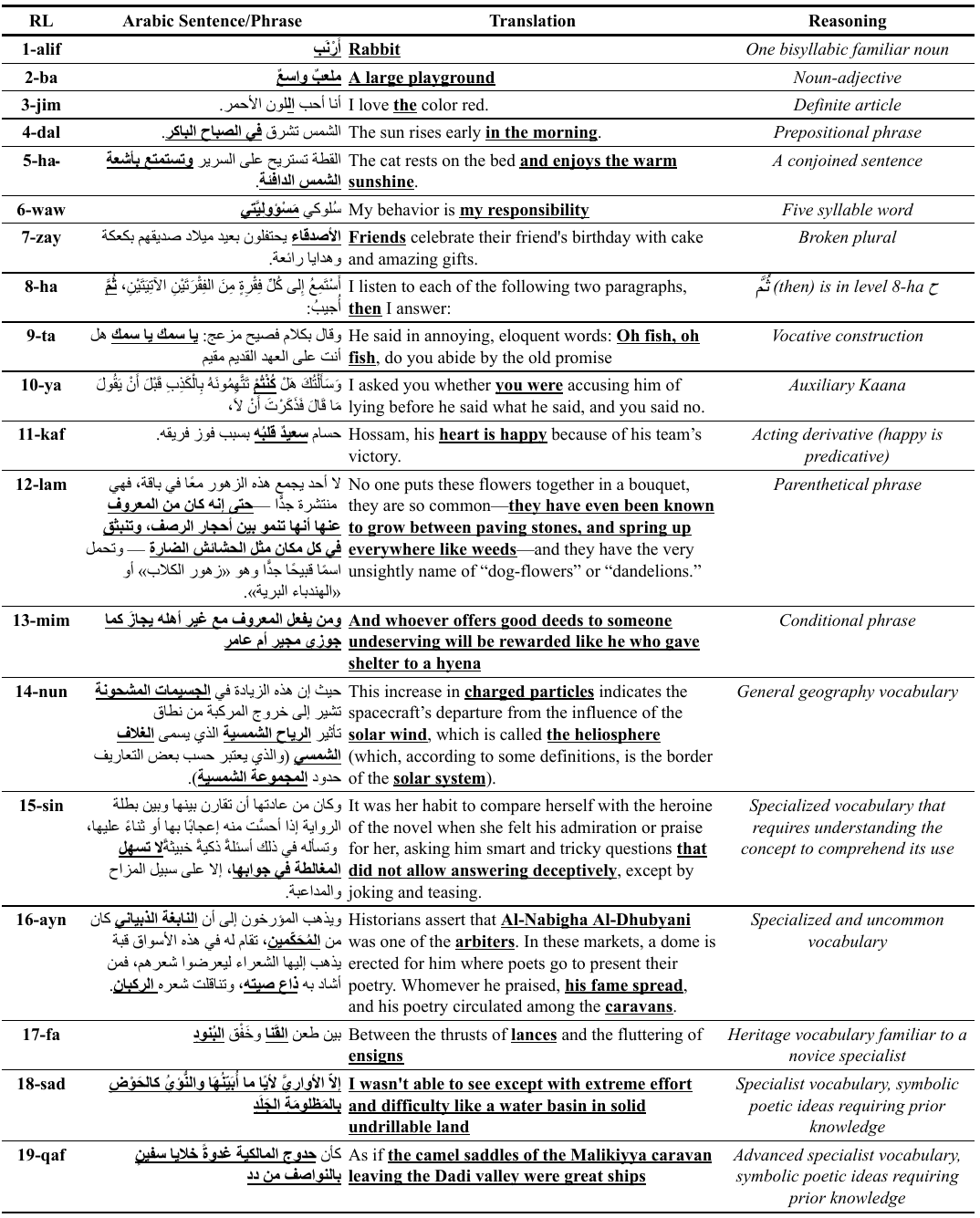}
    \caption{Representative subset of examples of the 19 {\barec} readability levels, with English translations, and readability level reasoning. Underlining is used to highlight the main keys that determined the level.  }
\label{tab:barec-examples}
\end{table*}


\subsection{Annotation Desiderata}
Our guidelines and annotation decisions follow several key principles. \textbf{Comprehensive Coverage} ensures representation across all 19 levels, from kindergarten to postgraduate, with finer distinctions at early stages. \textbf{Objective Standardization} defines levels using consistent linguistic and content-based criteria, avoiding overreliance on surface features like word or sentence length. \textbf{Bias Mitigation} promotes inclusivity across Arab world regions and cultural content. \textbf{Balanced Coverage} supports diversity in levels, genres, and topics, especially addressing material scarcity in areas like children's literature. \textbf{Quality Control} is maintained through trained annotators and regular checks for inter-annotator agreement and consistency. Finally, \textbf{Ethical Considerations} include respecting copyrights and fairly compensating annotators.

\subsection{Readability Levels}
\label{read-levels}
{\barec} readability annotation assigns one of 19 levels to each sentence in the corpus. 
We retain \newcite{Taha:2017:guidelines}'s 19-level naming system based on the Abjad order: {\BlevAlif},
{\BlevBa},
{\BlevJim}, ...
{\BlevQaf}, but extend and adjust the original guidelines, which were designed for book-level annotation to this task. 
The {\barec} pyramid (Figure~\ref{fig:barec-pyramid}) illustrates the scaffolding of these levels and their mapping to guidelines components, school grades, and three collapsed versions of level size 7, 5, and~3. All four level types (19-7-5-3) are fully aligned to allow easy mapping from fine-grained to coarse-grained levels, but manual annotation only happened on 19 levels. For example, level {\BlevKaf} maps to level \textbf{4}~(of~7), level \textbf{2}~(of~5) and level \textbf{1}~(of~3).
See Table~\ref{tab:barec-examples} for representative examples.

\subsection{Readability Annotation Principles}

\paragraph{Reading and Comprehension}
Readability reflects how easily independent readers can both read and comprehend a text without teacher or parent support. We focus on basic pronunciation (recovering lexical diacritics) and literal understanding, not on grammatical analysis or deep interpretation.  

 
\paragraph{Sentence-level Focus}
We assess readability at the sentence level, independent of broader context, source, or author intent. This deliberate choice avoids genre-based assumptions and enables fair, objective comparison across diverse texts. Mapping sentence-level judgments to larger units is left for future work.

\paragraph{Target Audience}
While religious content is part of basic public education in the Arab world, we make no assumptions about readers' religious backgrounds or prior knowledge. Readability is judged purely on linguistic and cognitive grounds. Our guidelines reflect Modern Standard Arabic (MSA) as used in Egypt, the Gulf, and the Levant, leaving variations in other regions for future work.

\paragraph{Readability Level Keys}
Annotators start from the lowest (easiest) level and raise it based on key features: lexical, morphological, syntactic, or semantic. See Sections~\ref{textkeys}~and \ref{process} below for details.

\paragraph{A Note on Arabic Diacritics}
While diacritics can aid comprehension, we assess readability without relying on them. This departs from \newcite{Taha:2017:guidelines}, who consider diacritics a key design feature in children’s books. In ambiguous cases, we choose the simpler meaning, e.g., \<هذه سلطة بدون خيار> \textit{h{\DHA}h slT{\TAMARBUTA} bdwn xyAr}\footnote{HSB transliteration \cite{Habash:2007:arabic-transliteration}.} is read as `a salad without cucumbers' not `an authority without choices'.


\subsection{Dimensions of Textual Features}\label{textkeys}
To determine the {\barec} level, we define six textual dimensions that identify \textit{key} features  necessary to unlock each level:

\paragraph{1. Number of Words}  
Counts unique printed words (ignoring punctuation and diacritics). Used only up to level {\BlevKaf} (max 20 words).  

\paragraph{2. Orthography \& Phonology}  
Focuses on word length (syllables) and letters like Hamzas. Final diacritics are ignored (words read in \textit{waqf}), e.g., \<أَرْنَبٌ> \textit{Âar.nabũ} `rabbit' has 2 syllables: \textit{ar-nab}.

\paragraph{3. Morphology}  
Covers derivation and inflection (tense, voice, number, etc.). Simpler forms appear at lower levels (e.g., present tense before past, singular before plural). Used up to level {\BlevMim}.

\paragraph{4. Syntactic Structures}  
Tracks sentence complexity, from single words (\BlevAlif) to complex constructions. Used up to level {\BlevSin}.

\paragraph{5. Vocabulary}  
Central at all levels. Overlapping dialect and MSA vocabulary appear at easier levels; technical terms are introduced at harder levels. Arabized foreign words are treated as part of the language, while non-Arabic script is excluded.

\paragraph{6. Ideas \& Content}  
Evaluates needed prior knowledge, symbolic unpacking, and conceptual linking. Levels progress from familiar to specialized knowledge and from literal to abstract ideas.
We recognize that such evaluations are complex and may vary subjectively among readers within the same age or education group.


\paragraph{Problems and Difficulties}
Annotators are instructed to report issues such as spelling errors, colloquial language, or sensitive topics. Difficulty is noted when annotations cannot be made due to conflicting guidelines.

The {\barec} pyramid (Figure~\ref{fig:barec-pyramid}) illustrates which aspects are used (broadly) for which levels. For example, spelling criteria are only used up to level {\BlevZay}, while syntax is used until level {\BlevSin}, and word count is not used beyond level {\BlevKaf}. 
A full set of examples with explanations of leveling choices is in Table~\ref{tab:barec-examples}. 
The \textbf{\textit{Annotation Cheat~Sheet}} used by the annotators in Arabic and its translation in English are included in Appendix~\ref{app:guidelines}. The full guidelines are publicly 
available.\textsuperscript{\ref{barec-site}}
For more on Arabic linguistic features, see \newcite{Habash:2010:introduction}.

\subsection{Annotation Process} \label{process}
\paragraph{Sentence Segmentation} Since our starting point is a text excerpt, typically a paragraph  or two ($\sim$500±200 words) from each source, we begin with sentence-level segmentation and initial text flagging. We followed the Arabic sentence segmentation guidelines by \newcite{habash-etal-2022-camel}.
\paragraph{Sentence Readability Annotation}
Each annotator is presented with a batch of 100 randomly selected sentences to annotate.
The annotation  was done through a simple Google Sheet interface (see Appendix~\ref{interface}), which provides details such as sentence word count, and the guidelines constraints for the selected level to provide feedback confirmation to the annotator.
The annotators are instructed to follow this procedure: \textbf{First} they read the sentence and make sure it has no flaws that can lead to excluding it. \textbf{Second}, they think about the meaning of the sentence noting any ambiguities due to diacritic absence or limited context, and consciously decide on the simpler reading in case of multiple readings. \textbf{Third}, they make an initial assessment of the lowest possible level based on word count.  \textbf{Fourth}, they look for specific phenomena that allow increasing the level to the highest possible.  
For example, the sixth sentence in Table~\ref{tab:barec-examples},
\<سلوكي مسؤوليتي> \textit{slwky msŵwlyty} `my behavior is my responsibility' has two words, which automatically sets it as level {\BlevBa} or higher. The presence of the first person pronominal clitic \<ي>+~\textit{+y} elevates the level to {\BlevJim}; however, the fact that the second word has five syllables raises the level further to {\BlevWaw}.  No other keys can take it higher. 

Annotation averaged 2.5 hours per 100-sentence batch (1.5 minutes per sentence), reflecting the careful and rigorous approach taken by annotators to ensure high-quality, consistent labeling across a diverse and challenging dataset.

 \subsection{Annotation Team}
The {\barec} annotation team included six native Arabic-speaking educators (A0-A5), most with advanced degrees in Arabic Literature or Linguistics. A0 had prior experience in computational linguistics annotation, while A1-A5 brought extensive expertise in readability assessment from the Taha/Arabi21 project. 
A0 handled sentence segmentation and initial text selection; and A5 led the annotation team in assigning readability labels.
Annotator profiles, covering demographic, educational, linguistic, and teaching backgrounds, are listed in Appendix~\ref{team}.

\subsection{Training and Quality Control}
Annotators A1-A5 received thorough training, including three shared pilot rounds that enabled in-depth discussion and refinement of the guidelines. 

To ensure consistency, the initial 10,658 sentences (Phase 1) were double-reviewed before annotating the full 69K (1M+ words). Inter-annotator agreement (IAA) was assessed on 19 blind batches (excluding pilots 1 and 2), followed by group unification to support quality control and prevent drift. Only unified labels appear in the official release.  The multiple IAA annotations will be released separately to support research on readability annotations.\textsuperscript{\ref{barec-site}} Details on IAA are in Section~\ref{sec:iaa-results}).

In total, the annotators labeled 92.6K sentences; 25\% were excluded from the final corpus: 3.3\% were problematic (typos and offensive topics), 11.5\% from early double annotations, and 10.3\% from IAA rounds (excluding unification).

\begin{table*}[h!]
\centering
\tabcolsep5pt
\small
\begin{tabular}{llrrrrr}
\toprule
\textbf{Category} & \textbf{Domain} & \textbf{Foundational} & \textbf{Advanced} & \textbf{Specialized} & \textbf{All} \\
\midrule
\multirow{4}{*}{\textbf{Documents}} 
& \textbf{Arts \& Humanities} & 562 (29\%) & 478 (25\%) & 327 (17\%) & \textbf{1,367 (71\%)} \\
& \textbf{Social Sciences}     & 44 (2\%)   & 168 (9\%)  & 163 (8\%)  & \textbf{375 (20\%)} \\
& \textbf{STEM}                & 27 (1\%)   & 85 (4\%)   & 68 (4\%)   & \textbf{180 (9\%)} \\
& \textbf{All}                & \textbf{633 (33\%)} & \textbf{731 (38\%)} & \textbf{558 (29\%)} & \textbf{1,922 (100\%)} \\
\midrule
\multirow{4}{*}{\textbf{Sentences}} 
& \textbf{Arts \& Humanities} & 24,978 (36\%) & 15,285 (22\%) & 10,179 (15\%) & \textbf{50,442 (73\%)} \\
& \textbf{Social Sciences}    & 2,270 (3\%)   & 5,463 (8\%)   & 6,586 (9\%)  & \textbf{14,319 (21\%)} \\
& \textbf{STEM}               & 533 (1\%)     & 1,948 (3\%)   & 2,199 (3\%)  & \textbf{4,680 (7\%)} \\
& \textbf{All}               & \textbf{27,781 (40\%)} & \textbf{22,696 (33\%)} & \textbf{18,964 (27\%)} & \textbf{69,441 (100\%)} \\
\midrule
\multirow{4}{*}{\textbf{Words}} 
& \textbf{Arts \& Humanities} & 274,497 (26\%) & 222,933 (21\%) & 155,565 (15\%) & \textbf{652,995 (63\%)} \\
& \textbf{Social Sciences}    & 26,692 (3\%)   & 110,226 (11\%) & 138,813 (13\%) & \textbf{275,731 (27\%)} \\
& \textbf{STEM}               & 12,879 (1\%)   & 48,501 (5\%)   & 49,265 (5\%)   & \textbf{110,645 (11\%)} \\
& \textbf{All}               & \textbf{314,068 (30\%)} & \textbf{381,660 (37\%)} & \textbf{343,643 (33\%)} & \textbf{1,039,371 (100\%)} \\
\bottomrule
\end{tabular}
\caption{ {\barec} corpus statistics in documents, sentences, and words, across domain and readership levels.}
\label{tab:corpus-stats}
\vspace{-5pt}
\end{table*}

\section{{\barec} Corpus}
\label{sec:corpus}

\subsection{Corpus Selection}
In the process of corpus selection, we aimed to cover a wide educational span as well as different domains and topics. We collected the corpus from $1,922$ documents, which we manually categorized into three domains: \textbf{Arts \& Humanities}, \textbf{Social Sciences}, and \textbf{STEM},\footnote{%
\textbf{Arts \& Humanities:} literature, philosophy, religion, education, and related news. 
\textbf{Social Sciences:} business, law, social studies, education, and related news. 
\textbf{STEM:} science, technology, engineering, math, education, and related news.}
and three readership groups: \textbf{Foundational}, \textbf{Advanced}, and \textbf{Specialized}.\footnote{%
\textbf{Foundational:} Learners up to 4th grade (age 10), focused on basic literacy skills. 
\textbf{Advanced:} Adult readers with average abilities, handling moderate complexity texts. 
\textbf{Specialized:} Advanced readers (typically 9th grade+), engaging with domain-specific texts.} Table~\ref{tab:corpus-stats} shows the distribution of the documents, sentences and words across domains and groups.
%
%
%
The corpus emphasizes educational coverage, with a higher-than-usual proportion of foundational-level texts. Domain variation reflects text availability and reader interest (more Arts \& Humanities, less STEM). Texts were sourced from  30 resources, all either public domain, within fair use, or used with permission. Some were selected due to existing annotations. Notably, 25\% of sentences came from new sources that were manually digitized. See Appendix~\ref{app:full-data} for resource details.

\begin{figure}[t]
\centering
  \includegraphics[width=\columnwidth]{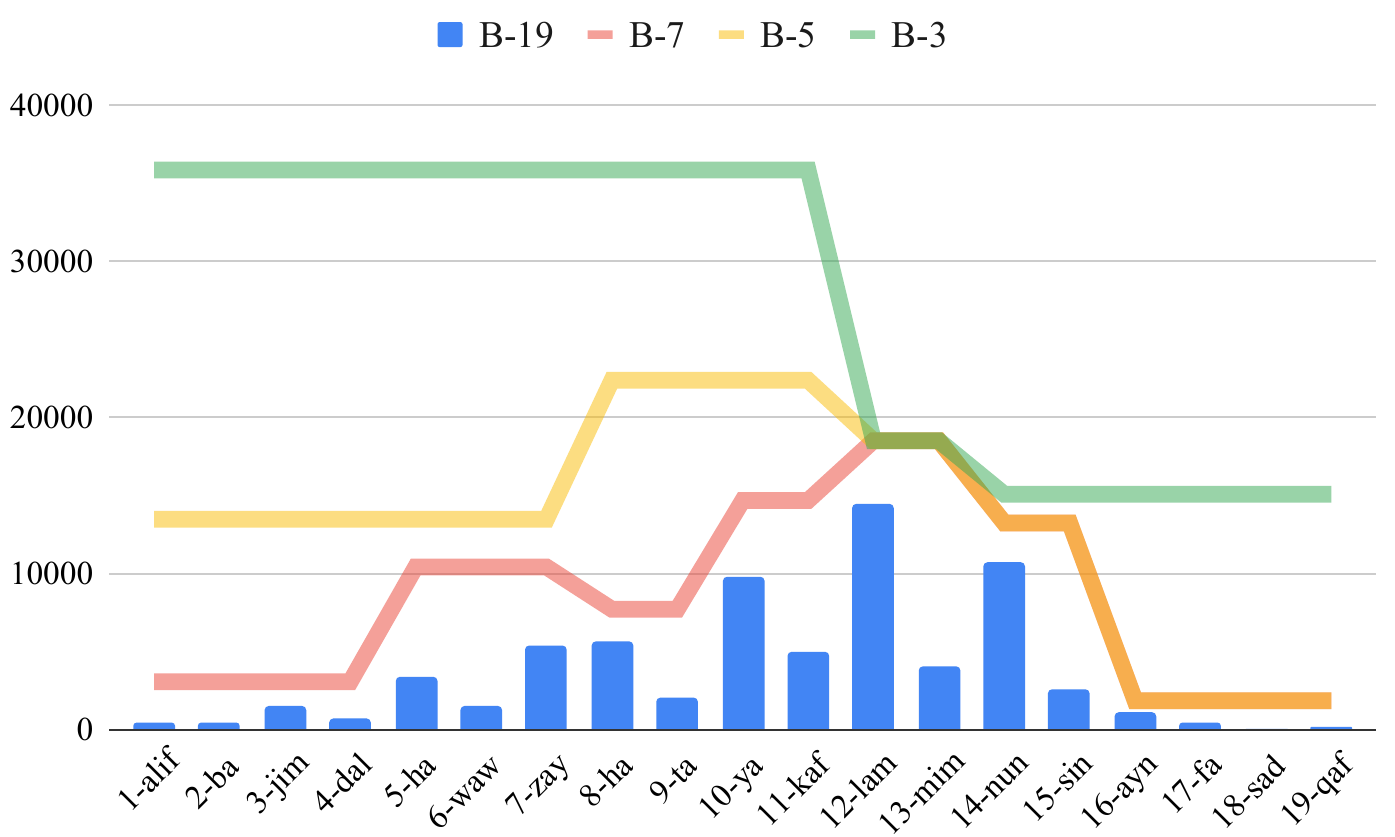}
    \caption{The distribution of sentences across {\barec}-19 levels (blue), and their mapping to coarser levels.}
\label{tab:readability-stats} 

\end{figure}
\begin{figure}[t]
\centering
  \includegraphics[width=\columnwidth]{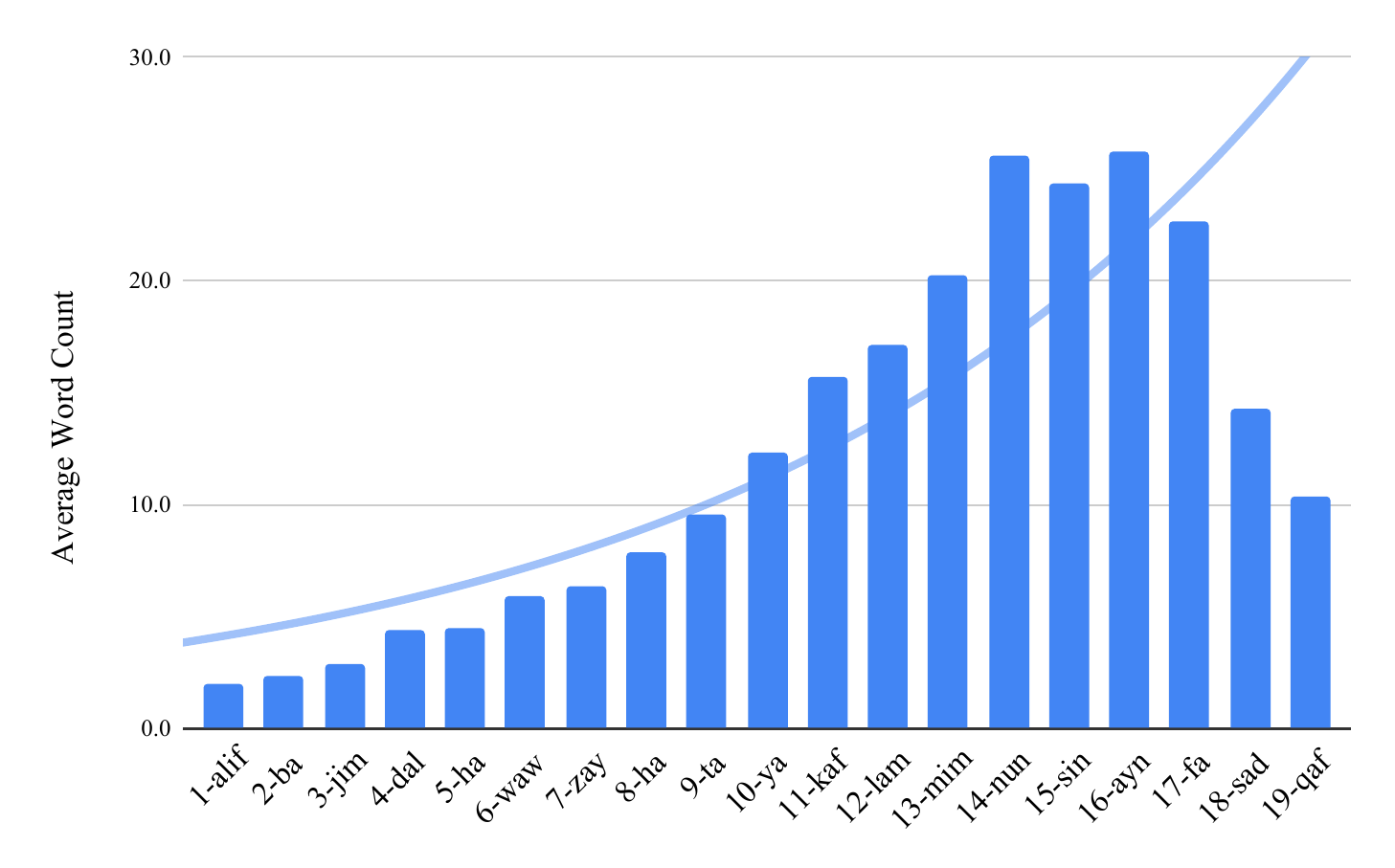}
    \caption{The average sentence word count  across {\barec}-19 levels, with trend line.}
\label{tab:readability-words}  
\end{figure}


\begin{figure*}[t]
\centering
  \includegraphics[width=\columnwidth]{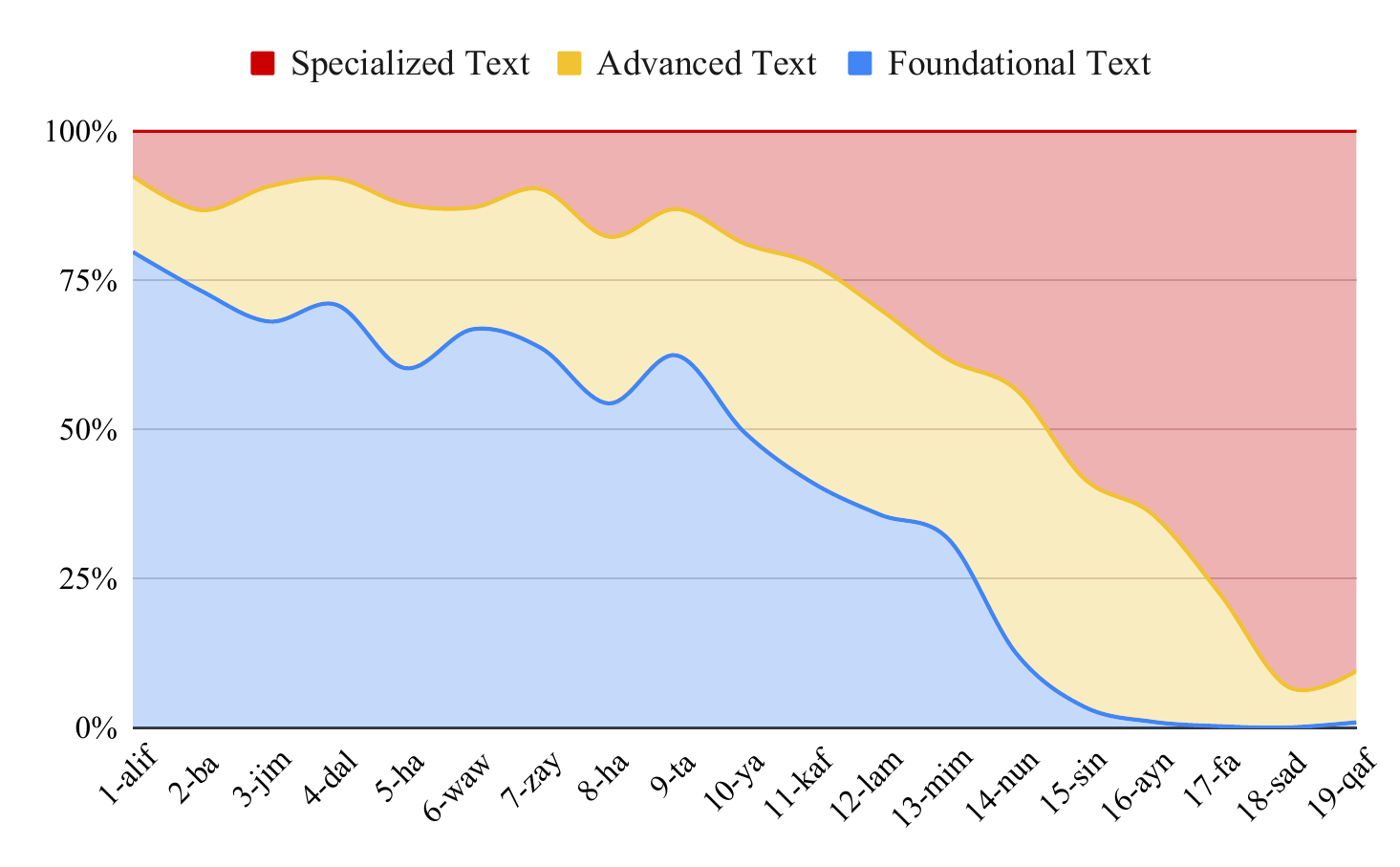} \includegraphics[width=\columnwidth]{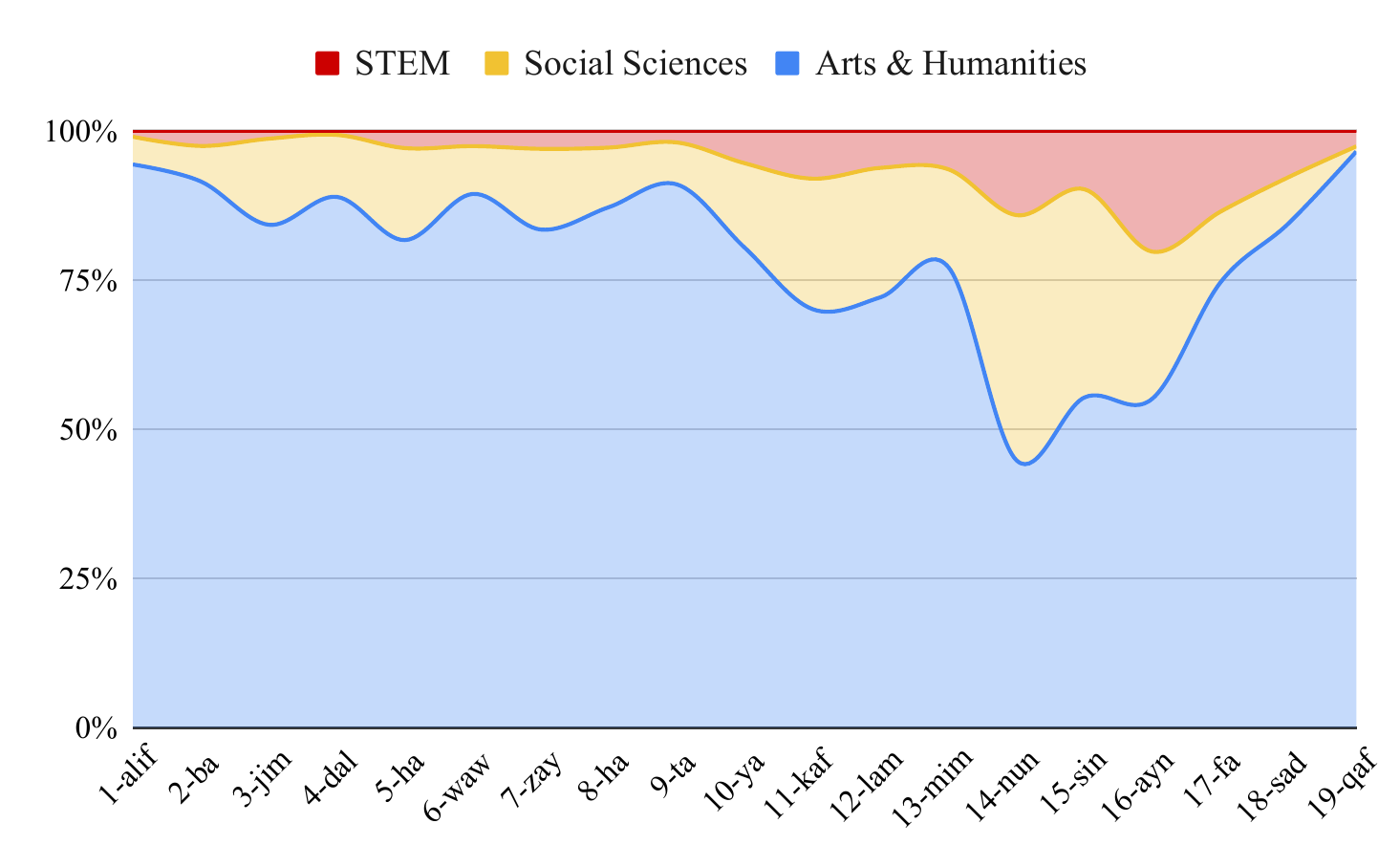}
    \caption{The relative distribution of readership groups and domains across {\barec} levels.}
\label{tab:readability-groups}
\end{figure*}



\subsection{Readability Statistics}

Figure \ref{tab:readability-stats} shows sentence distribution across {\barec}-19 levels and their mappings to coarser levels (7, 5, and 3). The distribution is uneven, with 63\% of sentences in the middle levels ({\BlevYa}$\sim$fourth grade to {\BlevNun}$\sim$ninth grade) reflecting  natural text complexity and real-world usage.

Figure \ref{tab:readability-words} shows average sentence length by level, which correlates strongly with readability (Pearson r=81\%). The drop at higher levels may result from shorter classical poetry lines.

Figure \ref{tab:readability-groups} shows \textit{relative} distribution of readership groups and domains across readability levels.  Foundational texts dominate lower levels and specialized texts higher ones. STEM and Social Science texts have a higher relative appearance in the upper mid levels.



\section{Evaluation and Analysis}
\label{sec:eval}

\subsection{Metrics}

We evaluate readability models and IAA using Accuracy, Adjacent Accuracy, Average Distance, and Quadratic Weighted Kappa (QWK), with QWK as our primary metric.

\paragraph{\textbf{Accuracy (Acc)}} The percentage of cases where the predicted class matches the reference class in the 19-level scheme (\textbf{Acc$^{19}$}), as well as three variants, \textbf{Acc$^{7}$}, \textbf{Acc$^{5}$}, and \textbf{Acc$^{3}$}, which collapse the 19-level scheme into 7, 5, and 3 levels, respectively (Section~\ref{read-levels}).

\paragraph{\textbf{Adjacent Accuracy (±1 Acc$^{19}$)}}
The proportion of predictions that are either exactly correct or off by at most one level.

\paragraph{\textbf{Average Distance (Dist)}}
The average absolute difference between two sets of labels. 
For example, the distance between {\BlevBa} and {\BlevDal} is 2.

\paragraph{\textbf{Quadratic Weighted Kappa (QWK)}}
An extension of Cohen’s Kappa \citep{Cohen:1968:weighted,2023.EDM-long-papers.9}, measuring agreement between predicted and true labels, with a quadratic penalty for larger misclassifications.

\subsection{Corpus Splits}

We split the corpus at the document level into \textbf{Train ($\sim$80\%)}, \textbf{Dev ($\sim$10\%)}, and \textbf{Test ($\sim$10\%)}. Sentences from IAA studies are distributed across splits.  For resources with existing splits, such as CamelTB \citep{habash-etal-2022-camel} and ReadMe++ \citep{naous-etal-2024-readme}, we adopted their original splits. Table~\ref{tab:corpus-splits} reports the splits by documents, sentences, and words. Due to IAA and external corpus constraints, final proportions slightly deviate from exact 80-10-10.
See Appendix~\ref{app:level-dist} for full and split readability level distributions.

\begin{table}[t]
\centering
\tabcolsep3pt
\small
\begin{tabular}{lrrr}
\toprule
\textbf{Split} & \textbf{\#Documents} & \textbf{\#Sentences} & \textbf{\#Words} \\
\midrule
Train & 1,518 (79\%) & 54,845 (79\%) & 832,743 (80\%) \\
Dev   & 194 (10\%)   & 7,310 (11\%)  & 101,364 (10\%) \\
Test  & 210 (11\%)   & 7,286 (10\%)  & 105,264 (10\%) \\
\midrule
\textbf{All} & 1,922 (100\%) & 69,441 (100\%) & 1,039,371 (100\%) \\
\bottomrule
\end{tabular}
\caption{{\barec} corpus splits.}
\label{tab:corpus-splits}
\end{table}


\begin{table}[t]
\centering
\small 
\tabcolsep2pt
\begin{tabular}{lccccc}
\toprule
\textbf{Stage} & \textbf{\#Sets} & \textbf{Distance} & \textbf{Acc$^{19}$} & \textbf{±1Acc$^{19}$}  & \textbf{QWK} \\
\midrule
Pilot 3     & 1  & 1.69 & 37.5\% & 58.5\% & 79.3\% \\
Phase 1   & 2  & 1.38 & 48.4\% & 64.4\% & 80.2\% \\
Phase 2A  & 6  & 1.21 & 49.4\% & 67.4\% & 72.4\% \\
Phase 2B  & 10 & 0.80 & 67.6\% & 78.3\% & 78.8\% \\
\midrule
Overall / Macro & 19 & 1.04 & 58.2\% & 72.3\% & 76.9\% \\

\midrule
Phase 2 / Macro & 16 & 0.96 & 60.8\% & 74.2\% & 76.4\% \\
Phase 2 / Micro & 16 & 0.95 & 61.1\% & 74.4\% & 81.8\% \\
\bottomrule
\end{tabular}
\caption{Average pairwise inter-annotator agreement (IAA) across different annotation stages. Macro/Micro indicate the form of averaging, over sets or sentences, respectively. Phase 2 = Phase 2A and 2B.}
\label{iaa-phases}
\end{table}

\subsection{Inter-Annotator Agreement (IAA)}
\label{sec:iaa-results}



\paragraph{Pairwise Agreement}
Table~\ref{iaa-phases} summarizes results for 19 IAA sets (excluding Pilots 1 and~2). We observe steady improvement from Pilot~3 to Phase~2B, with reduced distance and higher accuracy. The overall macro-average QWK is 76.9\%, indicating substantial agreement and suggesting that most disagreements are minor~\cite{Cohen:1968:weighted,2023.EDM-long-papers.9}. In Phase~2, the final and largest phase, the micro-average QWK rises to 81.8\%.

\begin{figure}[t]
\centering
  \includegraphics[width=\columnwidth]{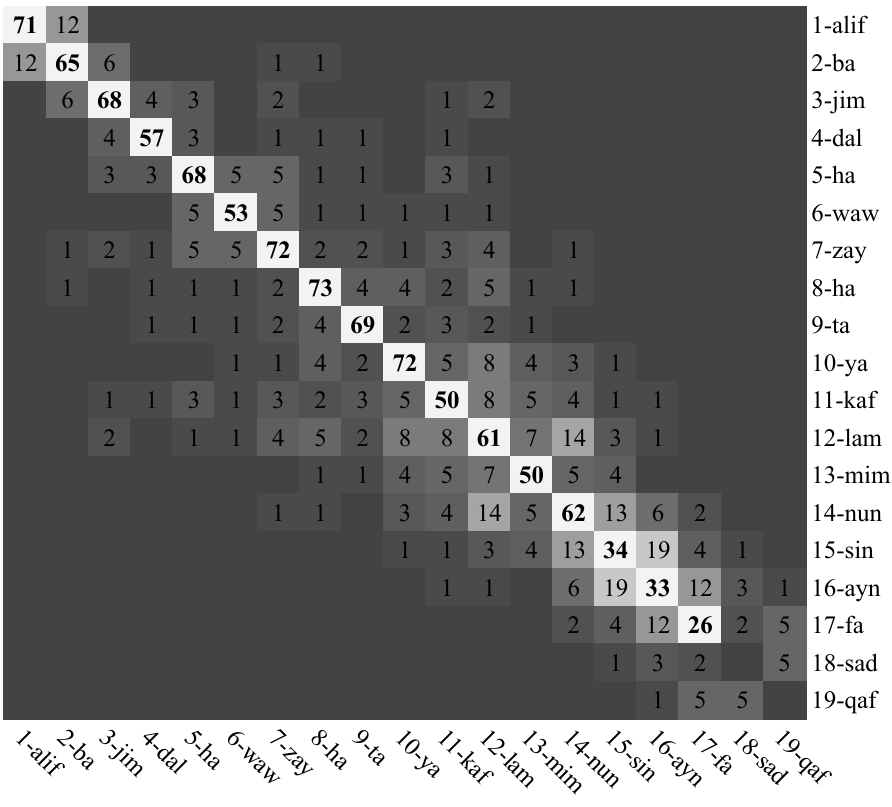}   
    \caption{Confusion matrix for annotator pairwise agreement on Phase 2 IAA sentences normalized as F-scores. }
\label{tab:iaa_confusion}
\end{figure}

Figure~\ref{tab:iaa_confusion} presents a confusion matrix of sentence-level pairwise agreements for Phase~2 IAA sentences, using F-scores to account for the unbalanced level distribution. The strong diagonal (exact matches) reflects a high degree of agreement, consistent with the overall IAA results. However, accuracy varies across levels, with more disagreement at the harder higher levels. This may stem from the guidelines  emphasizing vocabulary and content at the higher levels, features that are inherently more subjective than the textual feature cues used at lower levels.

\paragraph{Unification Agreement}
After each IAA study, annotators determined a unified readability level (UL) for each sentence. The UL falls within the Max-Min range of annotator labels 99.2\% of the time and matches one of the annotators 86.8\% of the time. Table~\ref{tab:iaa_results} compares the micro-average performance of annotators in Phase 2, using both pairwise comparisons and the comparison between the UL and the rounded average level (AL) of annotators' choices.
Table~\ref{tab:iaa_results} also presents the results mapped to lower granularity levels (7, 5 and 3).  
We observe that overall, the AL-UL distance is smaller than the average pairwise distance among the annotators, and that its ±1~Acc is much higher, which suggests the average (AL) is more often than not closer to UL than any pair of annotators are to each other. 
The comparison across granularity levels shows that although the absolute Distance decreases, its relative magnitude (compared to the label range) increases. As expected, both Acc and ±1~Acc are higher with coarser level groupings.
Appendix~\ref{tab:iaa_results_al-ul} presents the results for each annotator against UL.



\begin{table}[t]
\centering
\tabcolsep3pt
\small
\begin{tabular}{lrrrr}
\toprule
\textbf{} & \textbf{19 Level} & \textbf{7 Level} & \textbf{5 Level} & \textbf{3 Level} \\
\midrule
\textbf{Pairwise Distance}       & 0.95  & 0.39  & 0.30  & 0.23  \\
\multicolumn{1}{r}{\textit{Relative to Range}} & 5.0\% & 5.5\% & 6.0\% & 7.5\% \\
\textbf{Acc}            & 61.1\% & 73.1\% & 75.2\% & 80.0\% \\
\textbf{±1 Acc}         & 74.4\% & 92.0\% & 95.0\% & 97.3\% \\
\midrule
\textbf{AL-UL Distance} & 0.52  & 0.26  & 0.22  & 0.18  \\
\multicolumn{1}{r}{\textit{Relative to Range}} & 2.7\% & 3.7\% & 4.4\% & 5.9\% \\
\textbf{AL-UL Acc}      & 61.2\% & 75.5\% & 78.9\% & 82.9\% \\
\textbf{AL-UL ±1 Acc}   & 90.1\% & 98.5\% & 99.4\% & 99.5\% \\
\bottomrule
\end{tabular}
\caption{Comparison of pairwise agreement micro averages across level granularities for all Phase 2 IAA {sentences}. UL = Unified Label; AL = Average Label.}
\label{tab:iaa_results}
\end{table}

\begin{table*}[t!]
\centering
\setlength{\tabcolsep}{3pt} 

\small
\begin{tabular}{ccccccccp{5.5cm}}
\toprule
\textbf{Sentence (Arabic)} & \textbf{A1} & \textbf{A2} & \textbf{A3} & \textbf{A4} & \textbf{A5}    &\textbf{UL} & \textbf{MM} & \textbf{Comments} \\
\midrule
 \<أبي.. أبي..> 
& 2 & 2 & 2 & 3 & 3 & \textbf{3} & \textit{1} & \multirow{2}{=}{First person singular pronoun is level 3.} \\
 \textit{Dad .. Dad .. [lit. my father .. my father ..]} & \\\midrule
\addlinespace[0.5em]

\<احْتِضانُ الْأُمِّ لَهُم.>
& 9 & 12 & 5 & 5 & 5 & \textbf{5} & \textit{7} & \multirow{2}{=}{Disagreement over \textit{\<احتضان>} ‘embrace’: standard or dialect aligned.} \\
\textit{The mother's embrace for them.} & \\\midrule
\addlinespace[0.5em]


 \<أشعر بالتعب والجوع..>  
& 9 & 9 & 9 & 9 & 4 & \textbf{9} & \textit{5} & \multirow{2}{=}{Vocabulary describing emotions (level 9).} \\
\textit{I feel tired and hungry..}  & \\\midrule
\addlinespace[0.5em]

\<يتم ضمان حيادية الإدارة بموجب القانون.>
& 12 & 12 & 12 & 14 & 12 &  \textbf{12} & \textit{2} & \multirow{2}{=}{Disagreement over \textit{\<حيادية>} ‘neutrality’: general advanced or specialized.} \\
\textit{Administrative neutrality is guaranteed by law.} & \\ 

\bottomrule
\end{tabular}
\caption{Examples of Annotator Disagreements with Unified Levels (UL) and Max-Min Differences (MM)}
\label{tab:disagreements}
\end{table*}


\begin{table*}[t!]
\centering
\begin{tabular}{rccccccc}
\toprule
\textbf{Train} & \textbf{Distance} & \textbf{Acc$^{19}$} & \textbf{±1 Acc$^{19}$} & \textbf{QWK} & \textbf{Acc$^{7}$} & \textbf{Acc$^{5}$} & \textbf{Acc$^{3}$} \\
\midrule
12.5\%  & 1.35 & 45.0\% & 61.3\% & 77.2\% & 56.8\% & 63.0\% & 71.3\% \\
25.0\%  & 1.33 & 46.9\% & 63.0\% & 77.6\% & 58.8\% & 64.3\% & 72.3\% \\
50.0\%  & 1.16 & 52.4\% & 68.1\% & 80.7\% & 62.9\% & 67.6\% & 74.0\% \\
100.0\% & 1.09 & 55.8\% & 69.4\% & 81.0\% & 64.9\% & 69.1\% & 74.7\% \\
\bottomrule
\end{tabular}
\caption{Performance at different training data sizes across multiple evaluation metrics.}
\label{tab:train-size-performance}
\end{table*}

 \paragraph{Error analysis}

To better understand annotator disagreement, we manually analyzed 100 randomly selected sentences with divergent readability labels. Table~\ref{tab:disagreements} presents representative examples with explanations.  
We found that 25\% of disagreements were due to basic linguistic features (e.g., morphology, syntax, spelling), 12\% involved emotional or symbolic content, 18\% related to general advanced vocabulary, and 45\% stemmed from domain-specific terminology in STEM, Humanities, or Social Sciences. 
This suggests that specialized vocabulary is the leading source of inconsistency, often due to differing expectations about what counts as general versus domain-specific language, and how specialization is defined. Some variation also stems from subjective views on what an \textit{educated} Standard Arabic reader should know. In the future, we plan to develop readability lexicons to anchor our guidelines, building on efforts like the SAMER Lexicon \cite{al-khalil-etal-2020-large} and the Arabic Vocabulary Profile \cite{soliman2024creating}, but targeting 19 levels. 


\subsection{Automatic Readability Assessment}

To establish a baseline for sentence-level readability classification, we fine-tune AraBERTv02 \cite{antoun-etal-2020-arabert} using the Transformers library \cite{Wolf:2019:huggingfaces}.
Training is conducted on an NVIDIA V100 GPU for three epochs with a learning rate of \(5 \times 10^{-5}\), a batch size of 64, and a cross-entropy loss function for multi-class classification across 19 levels.
Table~\ref{tab:train-size-performance} presents the model's learning curve.
We evaluate performance using varying proportions of the training data: $1\over8$, $1\over4$, $1\over2$, and the full dataset.
As shown in the table, model performance improves consistently with larger training data.
Compared to the Phase 2 IAA micro averages (Table~\ref{iaa-phases}), the model's best Distance is 15.3\% higher, and its best Accuracy is 5.3\% absolute (8.7\% relative) lower. However, the QWK is only marginally lower by just 0.8\% absolute.
 

 For a more extensive discussion of the automatic annotation results, see \newcite{elmadani-etal-2025-readability}.

\section{Conclusions and Future Work} 

This paper presented the annotation guidelines of the  {Balanced Arabic Readability Evaluation Corpus (\textbf{{\barec}})}, a large-scale, finely annotated dataset for assessing Arabic text readability across 19 levels. With over 69K sentences and 1 million words, it is, to our knowledge, the largest Arabic readability corpus, covering diverse genres, topics, and audiences.  We report high inter-annotator agreement (QWK 81.8\% in Phase 2) that ensures reliable annotations.  Benchmark results across multiple classification granularities (19, 7, 5, and 3 levels) demonstrate both the difficulty and feasibility of automated Arabic readability prediction.

Looking ahead, we plan to expand the corpus by increasing its size and diversity to include more genres and topics. We also aim to add annotations for vocabulary leveling and syntactic treebanks to study the effect of vocabulary and syntax on readability. Future work will analyze readability variations across genres and topics. Additionally, we intend to integrate our tools into a system that assists children’s story writers in targeting specific reading levels.

The {\barec} dataset, its annotation guidelines, and benchmark results, are publicly available to support future research and educational applications in Arabic readability assessment.\textsuperscript{\ref{barec-site}}


\section*{Acknowledgments}
The {\barec} project is supported by the Abu Dhabi Arabic Language Centre (ALC) / Department of Culture and Tourism, UAE. 

We acknowledge the support of the High Performance Computing Center at New York University Abu Dhabi.

We are deeply grateful to our outstanding annotation team:  Mirvat Dawi, Reem Faraj, Rita Raad, Sawsan Tannir, and Adel Wizani, Samar Zeino, and Zeina Zeino. 

Special thanks go to Karin Aghadjanian, and Omar Al Ayyoubi of the ALC for their continued support.

We would also like to thank the Zayed University ZAI Arabic Language Research Center team, in particular Hamda Al-Hadhrami,  Maha Fatha, and Metha Talhak, for their valuable contributions to typing materials for the project.  We also acknowledge Ali Gomaa and his team for their additional support in this area.

Finally, we thank our colleagues at the New York University Abu Dhabi Computational Approaches to Modeling Language (CAMeL) Lab,  Muhammed Abu Odeh,  Bashar Alhafni, Ossama Obeid, and Mostafa Saeed, as well as Nour Rabih (Mohamed bin Zayed University of Artificial Intelligence)  for their helpful conversations and feedback.

\section*{Limitations}
One notable limitation is the inherent subjectivity associated with readability assessment, which may introduce variability in annotation decisions despite our best efforts to maintain consistency. Additionally, the current version of the corpus may not fully capture the diverse linguistic landscape of the Arab world. Finally, while our methodology strives for inclusivity, there may be biases or gaps in the corpus due to factors such as selection bias in the source materials or limitations in the annotation process. We acknowledge that readability measures   can be used with malicious intent to profile people; this is not our intention, and we discourage it.

\section*{Ethics Statement}

All data used in the corpus curation process are sourced responsibly and legally.
The annotation process is conducted with transparency and fairness, with multiple annotators involved to mitigate biases and ensure reliability. All annotators are paid fair wages for their contribution.  The corpus and associated guidelines are made openly accessible to promote transparency, reproducibility, and collaboration in Arabic language research.


\bibliography{custom,camel-bib-v3,anthology}

\onecolumn
\appendix
\section{{\barec} Annotation Guidelines Cheat Sheet and Annotation Interface}
\label{app:guidelines}
\subsection{Arabic Original}
\begin{center}
  \includegraphics[width=\textwidth]{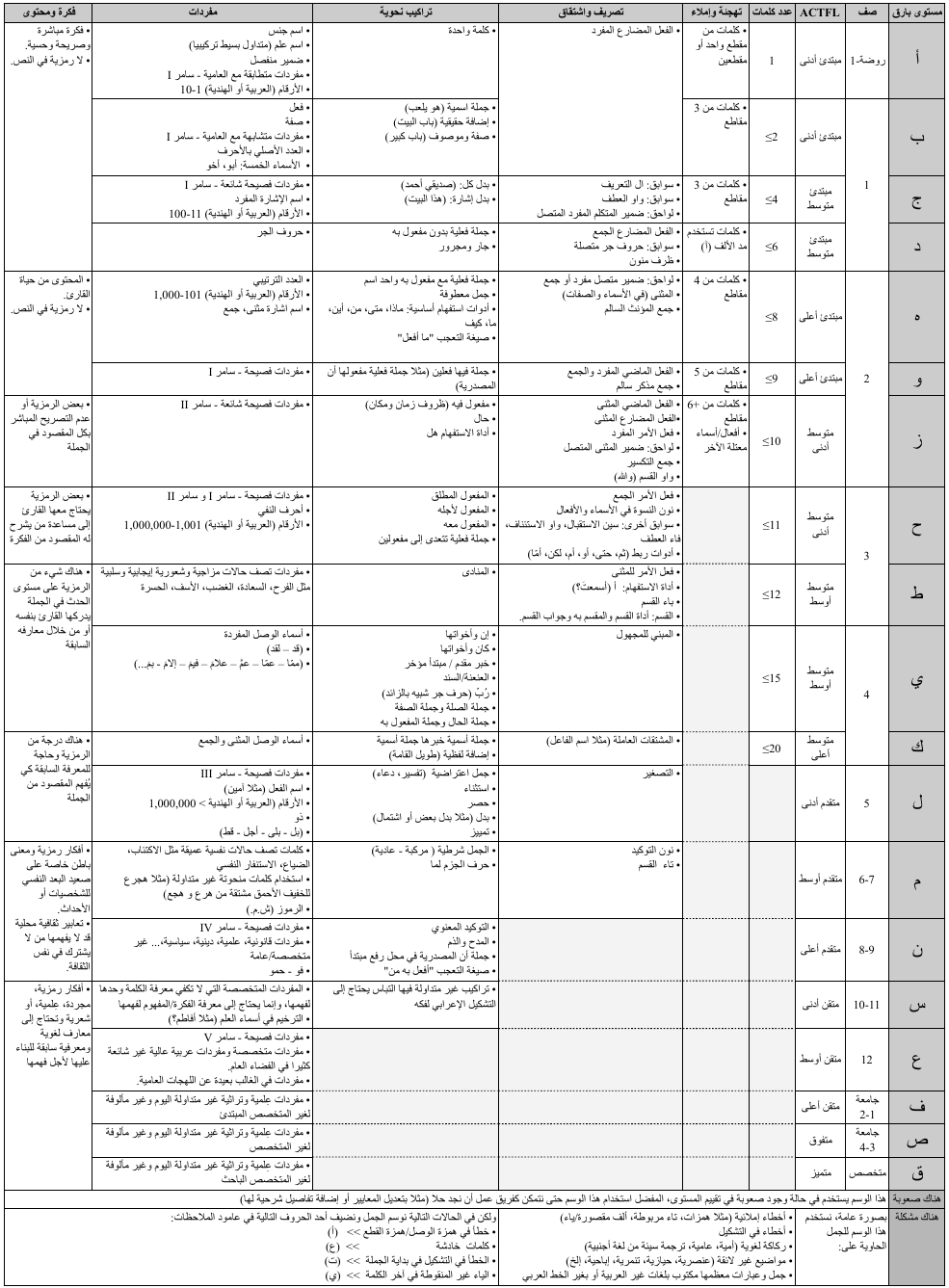}
  \end{center} 
\newpage 
\subsection{English Translation}
\label{fullexample}
\begin{center}
  \includegraphics[width=\textwidth]{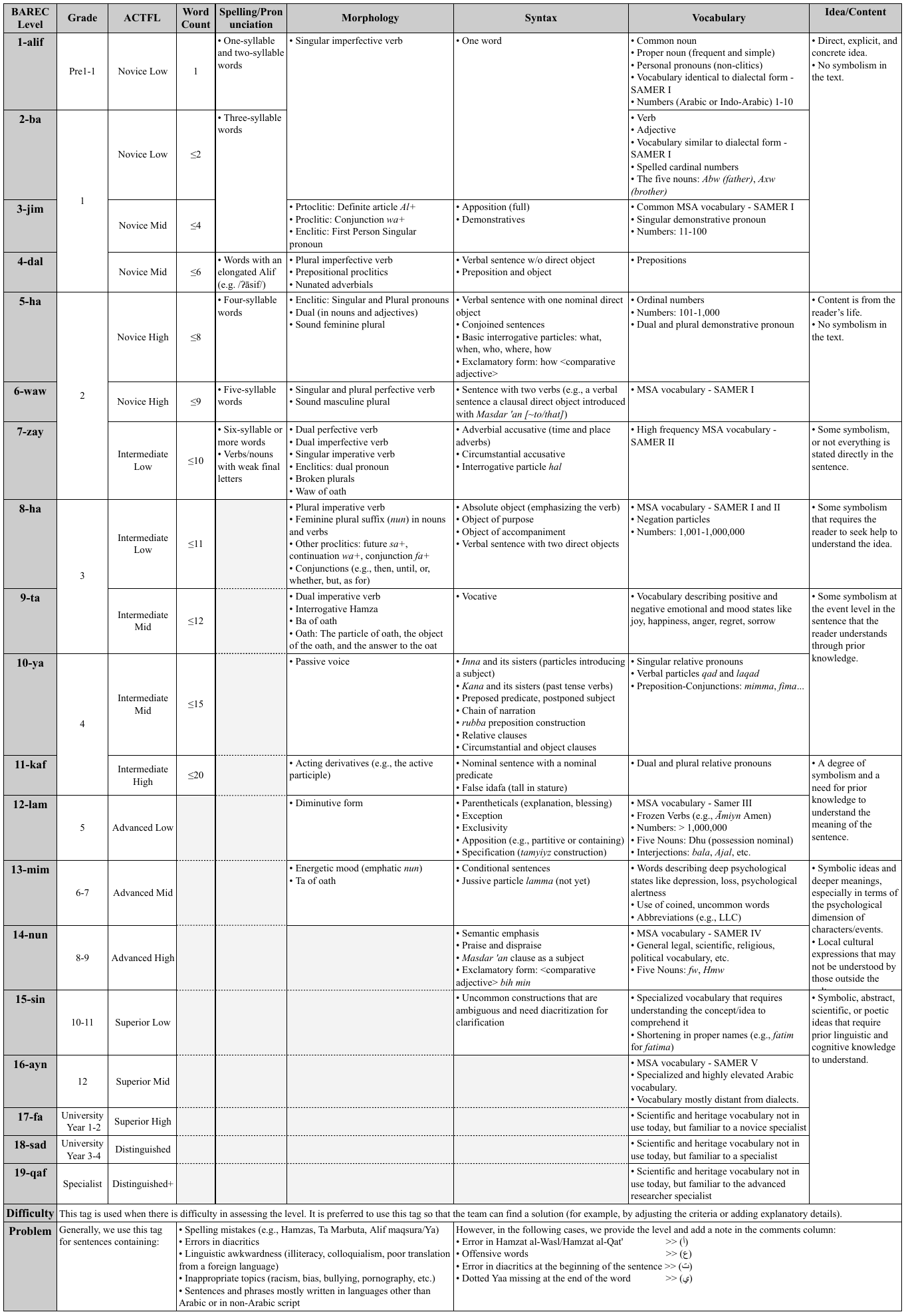}
  
\end{center}

\newpage 
\subsection{Annotation Interface}
\label{interface}
\begin{center}
  \includegraphics[width=\textwidth]{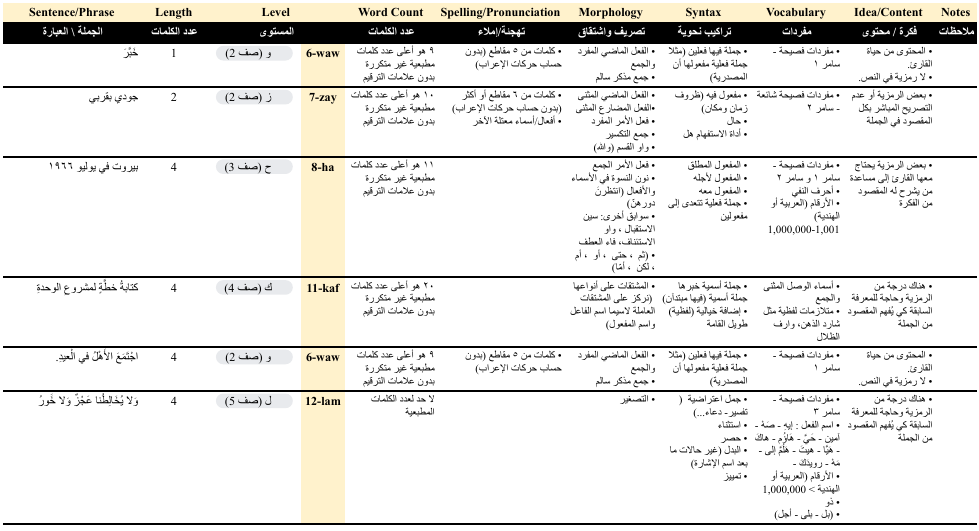}
\end{center} 
This is a screenshot of the Google Sheet interface used for annotation. 
The first two columns on the left are the sentence and its word count. The third column is the readability level which is selected by drop down menus. The fourth yellow column and the first yellow row are not part of the interface, we added them for the purpose of explaining the structure to readers of this paper who do not know Arabic. The next 6 columns automatically display the text features from the annotation guidelines to help the annotators confirm their choices. The last column is for extra notes such as flagging problematic sentences.

\subsection{Annotation Team}
\label{team}

\begin{table*}[h]
\setlength{\tabcolsep}{6pt} 

\centering
\begin{small}
\begin{tabular}{lcccccc}
\toprule

\textbf{ } & \textbf{A0$^P$} &\textbf{A1} & \textbf{A2} & \textbf{A3} & \textbf{A4} & \textbf{A5$^L$} \\
\midrule
\textbf{Native Language} & Arabic  & Arabic & Arabic & Arabic & Arabic & Arabic  \\
\textbf{Other Language} & En, Fr & En & En, Fr & En, Fr &En, Fr & En, Fr  \\
\textbf{Nationality} & Syrian & Lebanese & Lebanese & Lebanese & Lebanese & Lebanese \\
\textbf{Residence} & USA & Lebanon & Lebanon & Lebanon & UAE & Lebanon  \\
\textbf{Gender} & Female & Female & Female & Female & Female & Female  \\
\textbf{Background} & Muslim & Muslim & Muslim & Muslim & Christian & Muslim  \\
\textbf{Degree}  &   MA         & BA          &  BA       &  MA         &  MA   & B  MA   \\
\textbf{Major}   &  Applied Ling.  & Arabic Lit. & Geography & Arabic Lit. &  Arabic Lit.&  Arabic Lit. \\

\textbf{Experience} & CT, LA, RA & PT, LA & PT, LA & CT, LA  & CT, LA    &  CT, LA, RA   \\
\textbf{School} & Private & - & - & Public\&Private & Private & Public  \\
\textbf{Level} & University  & Elementary & Elementary & Secondary & Secondary & Secondary \\
\textbf{Students} & L2  & L1 & L1 & L1 & L1 & L1 \\
\textbf{Years} & 16 & 16 & 22 & 22 & 8 & 25  \\

\bottomrule
\end{tabular}%

\end{small}
\caption{Annotator background information. 
All have extensive linguistic annotation experience.
Certified Teacher (CT), Private Tutor (PT),
Linguistic Annotator (LA), Research Assistant (RA).
\textbf{A0$^P$} is the  preprocessing and segmentation lead; and \textbf{A5$^L$} is the readability annotation lead. 
}
\label{tab:annotator-background}
\end{table*}

\clearpage  
\subsection{Inter-Annotator Agreement between Annotator Labels and Unified Labels}
\label{tab:iaa_results_al-ul}
\begin{table*}[h!]
\centering
\begin{tabular}{lccccccc}
\toprule
      & \textbf{Acc$^{19}$} & \textbf{$\pm$1 Acc$^{19}$} & \textbf{Dist} & \textbf{QWK} & \textbf{Acc$^{7}$} & \textbf{Acc$^{5}$} & \textbf{Acc$^{3}$} \\
\midrule
\textbf{A1}    & 78.4\% & 89.0\% & 0.42 & 93.4\% & 85.3\% & 87.0\% & 89.7\% \\
\textbf{A2}    & 65.1\% & 76.4\% & 0.87 & 82.2\% & 71.6\% & 73.6\% & 79.3\% \\
\textbf{A3}    & 66.4\% & 78.4\% & 0.78 & 86.0\% & 73.7\% & 75.8\% & 79.0\% \\
\textbf{A4}    & 63.7\% & 76.6\% & 0.86 & 83.8\% & 71.8\% & 74.2\% & 79.5\% \\
\textbf{A5}    & 85.1\% & 91.2\% & 0.31 & 94.8\% & 89.2\% & 90.3\% & 92.9\% \\
\midrule
\textbf{Avg}   & 71.7\% & 82.3\% & 0.65 & 88.1\% & 78.4\% & 80.2\% & 84.1\% \\
\bottomrule
\end{tabular}
\caption{Inter-Annotator Agreement (IAA) results  comparing initial annotations by A1-A5 to unified labels (UL).}

\end{table*}

\section{{\barec} Corpus Level Distributions Across Splits} 
\label{app:level-dist}

\begin{table*}[h!]
\centering
\begin{tabular}{lrrrrrrrr}
\toprule
\textbf{Level} & \textbf{All} & \textbf{\%} & \textbf{Train} & \textbf{\%} & \textbf{Dev} & \textbf{\%} & \textbf{Test} & \textbf{\%} \\
\midrule
\BlevAlif   & 409   & 1\%  & 333   & 1\%  & 44    & 1\%  & 32    & 0\% \\
\BlevBa     & 437   & 1\%  & 333   & 1\%  & 68    & 1\%  & 36    & 0\% \\
\BlevJim    & 1,462 & 2\%  & 1,139 & 2\%  & 182   & 2\%  & 141   & 2\% \\
\BlevDal    & 751   & 1\%  & 587   & 1\%  & 78    & 1\%  & 86    & 1\% \\
\BlevHe     & 3,443 & 5\%  & 2,646 & 5\%  & 417   & 6\%  & 380   & 5\% \\
\BlevWaw    & 1,534 & 2\%  & 1,206 & 2\%  & 189   & 3\%  & 139   & 2\% \\
\BlevZay    & 5,438 & 8\%  & 4,152 & 8\%  & 701   & 10\% & 585   & 8\% \\
\BlevHa     & 5,683 & 8\%  & 4,529 & 8\%  & 613   & 8\%  & 541   & 7\% \\
\BlevTa     & 2,023 & 3\%  & 1,597 & 3\%  & 236   & 3\%  & 190   & 3\% \\
\BlevYa     & 9,763 & 14\% & 7,741 & 14\% & 1,012 & 14\% & 1,010 & 14\% \\
\BlevKaf    & 4,914 & 7\%  & 4,041 & 7\%  & 409   & 6\%  & 464   & 6\% \\
\BlevLam    & 14,471& 21\% & 11,318& 21\% & 1,491 & 20\% & 1,662 & 23\% \\
\BlevMim    & 4,039 & 6\%  & 3,252 & 6\%  & 349   & 5\%  & 438   & 6\% \\
\BlevNun    & 10,687& 15\% & 8,573 & 16\% & 1,072 & 15\% & 1,042 & 14\% \\
\BlevSin    & 2,547 & 4\%  & 2,016 & 4\%  & 258   & 4\%  & 273   & 4\% \\
\BlevAyn    & 1,141 & 2\%  & 866   & 2\%  & 114   & 2\%  & 161   & 2\% \\
\BlevFa     & 480   & 1\%  & 364   & 1\%  & 49    & 1\%  & 67    & 1\% \\
\BlevSad    & 103   & 0\%  & 67    & 0\%  & 13    & 0\%  & 23    & 0\% \\
\BlevQaf    & 116   & 0\%  & 85    & 0\%  & 15    & 0\%  & 16    & 0\% \\
\midrule
\textbf{Total} & 69,441 & 100\% & 54,845 & 100\% & 7,310 & 100\% & 7,286 & 100\% \\
\bottomrule
\end{tabular}
\caption{Distribution of sentence counts and percentages across readability levels and data splits.}
\label{tab-splits-stats}
\end{table*}


\twocolumn
\section{{\barec} Corpus Sources}
\label{app:full-data}
We present the corpus sources in groups of their general intended purpose.

Some datasets are chosen because they already have annotations available for other tasks. We list them independently of other collections they may be part of.  
For example, dependency treebank annotations exist  \cite{habash-etal-2022-camel} for the texts we included from the Arabian Nights, Quran and Hadith, Old and New Testament, Suspended Odes Odes, and Sara (which comes from Hindawi Foundation).


\subsection{Education}

\paragraph{Emarati Curriculum} The first five units of the UAE curriculum textbooks for the 12 grades in three subjects: Arabic language, social studies, Islamic studies \cite{Khalil:2018:leveled}.

\paragraph{ArabicMMLU} 6,205 question and answer pairs from the ArabicMMLU benchmark dataset \cite{koto-etal-2024-arabicmmlu}.

\paragraph{Zayed Arabic-English Bilingual Undergraduate Corpus (ZAEBUC)}
100 student-written articles from the Zayed University Arabic-English Bilingual Undergraduate Corpus \cite{habash-palfreyman-2022-zaebuc}.

\paragraph{Arabic Learner Corpus (ALC)}
16 L2 articles from the Arabic Learner Corpus \citep{phdthesis}.

\paragraph{Basic Travel Expressions Corpus (BTEC)}
20 documents from the MSA translation of the Basic Traveling Expression Corpus \cite{eck-hori-2005-overview,takezawa-etal-2007-multilingual,bouamor-etal-2018-madar}.

\paragraph{Collection of Children poems} Example of the included poems: My language sings (\<لغتي تغني>), and Poetry and news (\<أشعار وأخبار>) \cite{kashkol, poetry-and-news}.

\paragraph{ChatGPT} To add more children's materials, we ask Chatgpt to generate 200 sentences ranging from 2 to 4 words per sentence, 150 sentences ranging from 5 to 7 words per sentence and 100 sentences ranging from 8 to 10 words per sentence.\footnote{\url{https://chatgpt.com/}} Not all sentences generated by ChatGPT were correct. We discarded some sentences that were flagged by the annotators. Table~\ref{tab:chatgpt} shows the prompts and the percentage of discarded sentences for each prompt.

\begin{table*}[t!]
\centering
  \includegraphics[width=1.8\columnwidth]{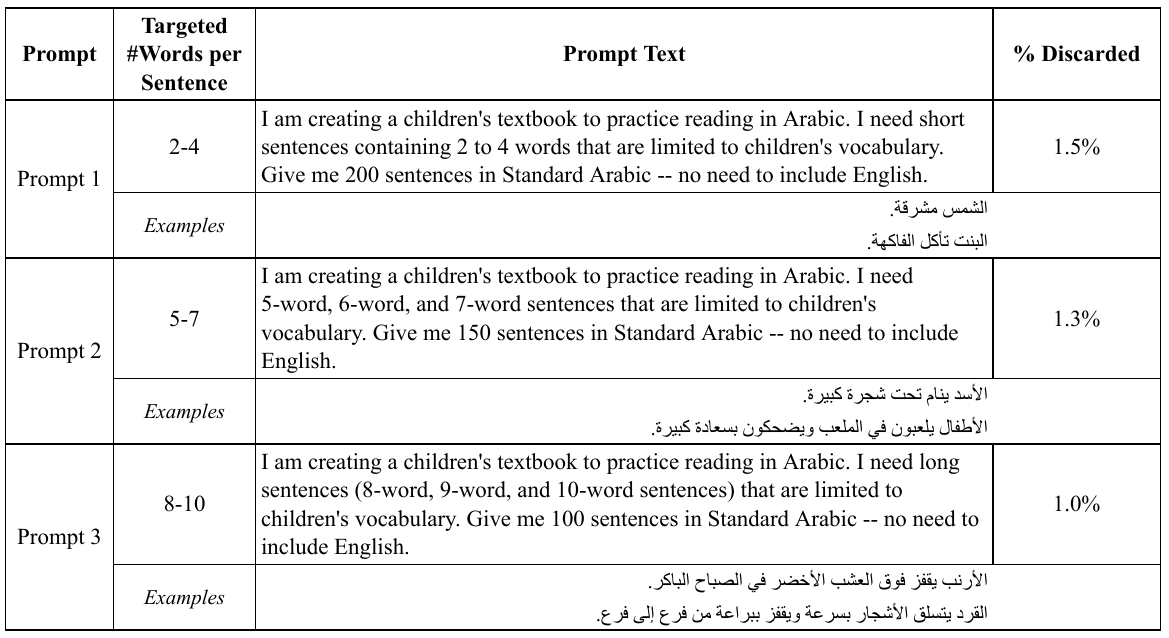}
    \caption{ChatGPT Prompts. \% Discarded is the percentage of discarded sentences due to grammatical errors.}
\label{tab:chatgpt}
\end{table*}

\subsection{Literature}

\paragraph{Hindawi} A subset of 264 books extracted from the Hindawi Foundation website across different different genres.\footnote{\url{https://www.hindawi.org/books/categories/}}

\paragraph{Kalima}
The first 500 words of 62 books from Kalima project.\footnote{\url{https://alc.ae/publications/kalima/}}

\paragraph{Green Library}
58 manually typed books from the Green Library.\footnote{\url{https://archive.org/details/201409_201409}}

\paragraph{Arabian Nights} The openings and endings of the opening narrative and the first eight nights from the Arabian Nights \cite{ArabianNights}. We extracted the text from an online forum.\footnote{\url{http://al-nada.eb2a.com/1000lela\&lela/}}

\paragraph{Hayy ibn Yaqdhan}  A subset of the philosophical novel and allegorical tale written by Ibn Tufail \cite{tufail:hayy}.
We extracted the text from the Hindawi Foundation website.\footnote{\url{https://www.hindawi.org/books/90463596/}}

\paragraph{Sara} The first 1000 words of {\it Sara}, a novel by Al-Akkad first published in 1938 \cite{akkad:sarah}. We extracted the text from the Hindawi Foundation website.\footnote{\url{https://www.hindawi.org/books/72707304/}}

\paragraph{The Suspended Odes (Odes)}  The ten most celebrated poems from Pre-Islamic Arabia (\<المعلقات> Mu’allaqat).
All texts were extracted from Wikipedia.\footnote{\url{https://ar.wikipedia.org/wiki/}\<المعلقات>}

\subsection{Media}

\paragraph{Majed}
10 manually typed editions of Majed magazine for children from 1983 to 2019.\footnote{\url{https://archive.org/details/majid_magazine}}

\paragraph{ReadMe++}
The Arabic split of the ReadMe++ dataset \cite{naous-etal-2024-readme}.

\paragraph{Spacetoon Songs}
The opening songs of 53 animated children series from Spacetoon channel.

\paragraph{Subtitles} A subset of the Arabic side of the OpenSubtitles 
dataset \cite{Lison:2016:opensubtitles2016}.

\paragraph{WikiNews} 62 Arabic WikiNews articles covering politics, economics,
health, science and technology, sports, arts, and culture \cite{Abdelali:2016:farasa}.

\subsection{References} 
\paragraph{Wikipedia} A subset of 168 Arabic wikipedia articles covering Culture, Figures, Geography, History, Mathematics, Sciences, Society, Philosophy, Religions and Technologies.\footnote{\url{https://ar.wikipedia.org/}}

\paragraph{Constitutions}
The first 2000 words of the Arabic constitutions from 16 Arabic speaking countries, collected from MCWC dataset \cite{el-haj-ezzini-2024-multilingual}.

\paragraph{UN} The Arabic translation of the Universal Declaration of Human Rights.\footnote{\url{https://www.un.org/ar/about-us/universal-declaration-of-human-rights}}

\subsection{Religion}

\paragraph{Old Testament} The first 20 chapters of the Book of  Genesis \cite{arabicOldTestament}.\footnote{\url{https://www.arabicbible.com/}\label{biblefoot}}

\paragraph{New Testament} The first 16 chapters of the Book of Matthew \cite{arabicNewTestament}.$^{\ref{biblefoot}}$ 

\paragraph{Quran} The first three Surahs and the last 14 Surahs from the Holy Quran. We selected the text from the Quran Corpus Project \cite{Dukes:2013:supervised}.\footnote{\url{https://corpus.quran.com/}}

\paragraph{Hadith} The first 75 Hadiths from Sahih Bukhari \cite{bukhari}.  We selected the text from the LK Hadith Corpus\footnote{\url{https://github.com/ShathaTm/LK-Hadith-Corpus}} \cite{Altammami:2019:Arabic}.

\end{document}